\documentclass[runningheads]{llncs}
\DeclareUnicodeCharacter{00B1}{\pm}
\usepackage{graphicx}
\usepackage[misc]{ifsym}
\usepackage{amsmath,amssymb}
\usepackage{mathrsfs}
\usepackage[ruled,linesnumbered]{algorithm2e}
\usepackage{multirow}
\usepackage{booktabs}

\begin{document}
\title{Every Corporation Owns Its Structure: Corporate Credit Ratings via Graph Neural Networks}
%
%
\author{Bojing Feng\inst{1,2}\textsuperscript{(\Letter)} \and
Haonan Xu\inst{1,2} \and
Wenfang Xue\inst{1,2,*}\and
Bindang Xue\inst{3}}
\titlerunning{Corporate Credit Ratings via Graph Neural Networks}
\authorrunning{Bojing Feng et al.}
%
\institute{Center for Research on Intelligent Perception and Computing, National Laboratory of Pattern Recognition, Institute of Automation, Chinese Academy of Science \and School of Artificial Intelligence, University of Chinese Academic of Science
\and
School of Astronautics, Beihang University \\
\email{bojing.feng,@cripac.ia.ac.cn},
\email{xuhaonan2020@ia.ac.cn}\\
\email{wenfang.xue@ia.ac.cn},
\email{xuebd@buaa.edu.cn}}
\maketitle              
\begin{abstract}
Credit rating is an analysis of the credit risks associated with a corporation, which reflects the level of the riskiness and reliability in investing, and plays a vital role in financial risk. There have emerged many studies that implement machine learning and deep learning techniques which are based on vector space to deal with corporate credit rating. Recently, considering the relations among enterprises such as loan guarantee network, some graph-based models are applied in this field with the advent of graph neural networks. But these existing models build networks between corporations without taking the internal feature interactions into account. In this paper, to overcome such problems, we propose a novel model, Corporate Credit Rating via Graph Neural Networks, CCR-GNN for brevity. We firstly construct individual graphs for each corporation based on self-outer product and then use GNN to model the feature interaction explicitly, which includes both local and global information. Extensive experiments conducted on the Chinese public-listed corporate rating dataset, prove that CCR-GNN outperforms the state-of-the-art methods consistently.

\end{abstract}

\section{Introduction}\label{section:introduction}
Nowadays, credit rating is fundamental for helping financial institutions to know companies well so as to mitigate credit risks~\cite{golbayani2020application}. It is an indication of the level of the risk in investing with the corporation and represents the likelihood that the corporation pays its financial obligations on time. Therefore, it is of great importance to model the profile of the corporation~\cite{golbayani2020comparative} to predict the credit rating level. However, this assessment process is usually very expensive and complicated, which often takes months with many experts involved to analyze all kinds of variables, which reflect the reliability of a corporation. One way to deal with this problem may be to build a model based on historical financial information~\cite{provenzano2020machine} of the corporation.

The banking industry has developed some credit risk models since the middle of the twentieth century. The risk rating is also the main business of thousands of worldwide corporations, including dozens of public companies~\cite{bravo2015improving}. Due to the highly practical value, many kinds of credit rating models have been developed. Traditionally, the credit models are proposed by logistic regression algorithms with the temporal credit rating as well as aggregated financial information.

Nowadays, machine learning and deep learning models have shown their power in a range of applications including financial fields. Parisa et al \cite{golbayani2020comparative} apply four machine learning techniques (Bagged Decision Trees, Random Forest, Support and Multilayer Perceptron) to predict corporate credit rating. Recently,~\cite{provenzano2020machine} builds a stack of machine learning models aiming at composing a state-of-the-art credit rating system. In the work~\cite{golbayani2020application}, they analyze the performance of four neural network architectures including MLP, CNN, CNN2D, LSTM in predicting corporate credit rating as issued by Standard and Poor’s.

With the advent of graph neural networks, some graph-based models~\cite{barja2019assessing,bruss2019deeptrax,cheng2020contagious,cheng2019risk,cheng2020spatio,cheng2019dynamic} are built based on the loan guarantee network. The corporations guarantee each other and form complex loan networks to receive loans from banks during the economic expansion stage.

Although these approaches are widely used and useful, we also observe that they have some limitations for corporate credit rating. Firstly, the existing deep learning models require extensive feature construction and specific background knowledge to design representative features. These features need to be aggregated from financial data, which costs most of the time. What’s more, the graph-based models usually regard a single corporation as a node in graph and build the relations between them, which neglects the feature interaction in a single corporation.

To overcome the limitations mentioned above, we propose a novel method, Corporate Credit Ratings via Graph Neural Networks, CCR-GNN for brevity. In contrast to previous graph-based approaches with global structure, we look at this problem from a new perspective. We regard the corporation as a graph instead of a node, which can depict the detailed feature-feature relations. The individual graphs are built by applying the corporation-to-graph method, which models the relations between features. Then the information of feature nodes is propagated to their neighbors via the Graph Neural Networks, graph attention network specifically, which takes advantage of attention mechanism. We also conduct extensive experiments to examine the advantages of our approach against baselines. 

To sum up, the main contributions of this work are summarized as follows: 
\begin{itemize}
    \item[$\bullet$]A new method named corporation-to-graph is developed to explore the relations between features.
    \item[$\bullet$]A new graph neural network for corporate credit rating is proposed, where each corporation is an individual graph and feature level interactions can be learned. To the best of our knowledge, this is the first work that applies graph neural networks into corporate credit rating with a graph-level perspective. It opens new doors to explore the advanced GNNs methods for corporate credit rating.
    \item[$\bullet$]We demonstrate that our approach outperforms state-of-the-art methods experimentally.
\end{itemize}

This paper is organized as follows. Related works about the researches are introduced in Section \ref{section:RelateWorks}. Section \ref{section:CCRGNN} will present the proposed CCR-GNN. The experiment results of CCR-CNN on real-world data will be presented in Section \ref{section:Experiments}. Finally, the conclusion is introduced in Section \ref{section:conclusion}.

\section{Related Works}\label{section:RelateWorks}
In this section, we review some related works on credit rating, including statistical models, machine learning models and hybrid models. Then we introduce the graph neural networks and graph-based models.

~\\
\noindent\textbf{Statistical Models.}
Researchers apply some traditional statistical models such as logistic regression. In bank credit rating, Gogas et al.~\cite{gogas2014forecasting} used an ordered probit regression model. Recently, the work ~\cite{petropoulos2016novel} proposed a model based on Student’s-t Hidden Markov Models (SHMMs) to investigate the firm-related data.

~\\
\noindent\textbf{Machine Learning Models.}
Nowadays, machine learning techniques are used to predict corporate ratings. The work \cite{kim2005predicting} implemented adaptive learning networks (ALN) on both financial data and non-financial data to predict S\&P credit ratings. Cao et al. \cite{cao2006bond} studied support vector machine methods on US companies.

~\\
\noindent\textbf{Hybrid Models.}
In addition these methods above, some researchers proposed hybrid models by mixing these techniques up. Yeh et al.~\cite{yeh2012hybrid} combined random forest feature selection with rough set theory (RST) and SVM. Pai et al. \cite{pai2015credit} built the Decision Tree Support Vector Machine (DTSVM) integrated TST. The work \cite{wu2012credit} proposed an enhanced decision support model that used the relevance vector machine and decision tree.

~\\
\noindent\textbf{Graph Neural Networks.}
Nowadays, neural networks have been developed for graph-structured data, such as social network and citation network. DeepWalk \cite{perozzi2014deepwalk} is designed to learn representations of graph nodes by random walk. Follow this work, unsupervised algorithm LINE \cite{tang2015line} and node2vec \cite{grover2016node2vec} are proposed. Besides, neural networks are applied in graph-structured data. An approach \cite{kipf2016semi} uses the convolutional architecture by a localized approximation of spectral graph convolutions. Then GAT \cite{velivckovic2017graph} utilizes the attention mechanism to improve the power of graph model.

~\\
\noindent\textbf{Graph-based Models.}
Due to the advance of GNNs, many researchers proposed graph-based models to apply in financial risk. Barja et al.~\cite{barja2019assessing} extracted a financial network from customer-supplier transactions among more than 140,000 companies, and their economic flows. In the work~\cite{cheng2020contagious}, they employed the temporal inter-chain attention network on graph-structured loan behavior data. Cheng et al.~\cite{cheng2019risk} proposed HGAR to learn the embedding of guarantee networks. The work developed DeepTrax~\cite{bruss2019deeptrax} in order to learn embeddings of account and merchant entities. Recently, work \cite{cheng2020spatio} combined the spatio-temporal information for credit fraud detection.
\section{The Proposed Method: CCR-GNN}\label{section:CCRGNN}
In this section, we introduce the proposed CCR-GNN which applies graph neural networks into corporate credit rating with the graph level. We formulate the problem at first, then give an overview of the whole CCR-GNN, and finally describe the three layers of the model: corporation to graph layer (C2GL), graph feature interaction layer (GFIL) and credit rating layer (CRL).

\begin{figure}[htbp]  
\centering  
\includegraphics[scale=0.145]{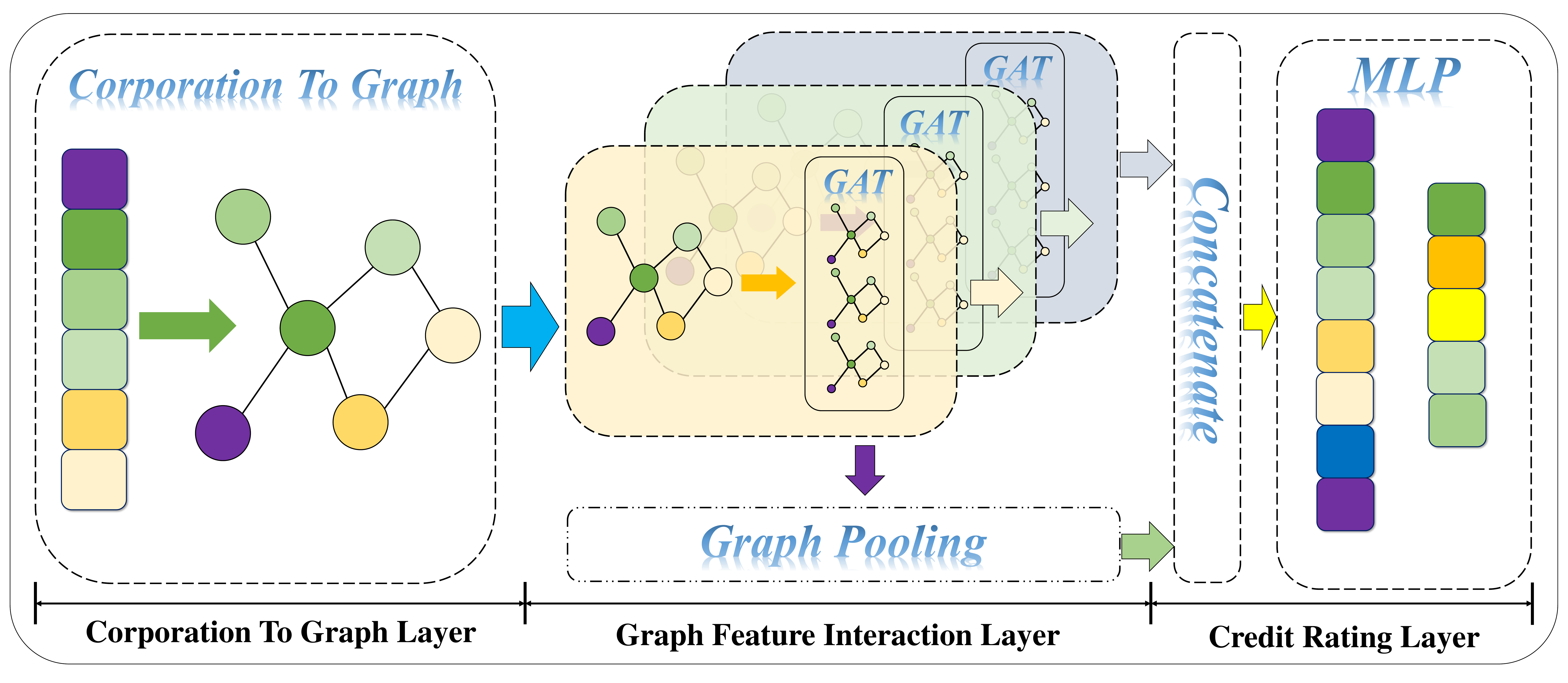}  
\caption{An Example of the architecture of our CCR-GNN model which has three Graph ATtention layers (GAT).} 
\label{CCRGNN}
\end{figure}

\subsection{Problem Formulation}
The credit rating system aims to predict which credit level of the corporation will belong to. Here we give a formulation of this problem as below.

In credit rating, let $C=\{c_1,c_2,\cdots,c_n\}$ denotes the set consisting of all unique corporations, $n$ is the number of corporations.$c\in \mathbb{R}^d$ Every corporation has a corresponding label which represents its credit level. Let $Y=\{y_1,y_2,\cdots,y_m\}$ denotes the set of the labels, and $m$ represents the number of unique labels. The goal of the credit rating system is to predict the corporate label according to its profile $c$. Under the credit rating model, for each corporation $c$, model outputs probabilities $\hat{y}$ for all labels, where an element value of vector $\hat{y}$ is the score of corresponding label. Finally, it will predict the label with the max score.

\subsection{Architecture Overview}
Fig. \ref{CCRGNN} illustrates the end-to-end CCR-GNN model. It is composed of three functional layers:  corporation to graph layer, graph feature interaction layer and credit rating layer. Firstly, every corporation is mapped into a graph-structured representation through corporation-to-graph to model corporation feature relations. Then the features interact with each other by graph attention network. Finally, credit rating layer outputs the label scores by utilizing the information provided by the before local and global information.

\subsection{Corporation To Graph Layer(C2GL)}
\begin{figure}[htbp]  
\centering  
\includegraphics[scale=0.145]{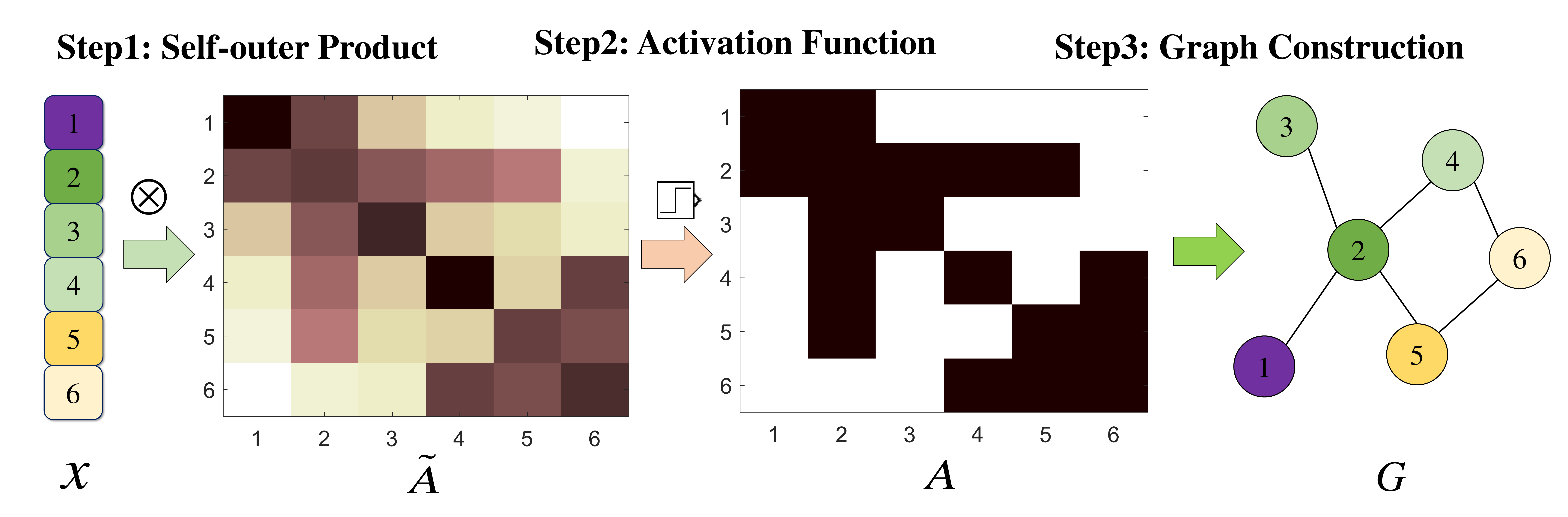}  
\caption{Corporation To Graph (C2G)} 
\label{C2G}
\end{figure}

For the convenience of formulation, we use $c$ to denote any corporation. It includes its financial data and non-financial data which describe the corporate profile. For numerical data, we can use it directly. However, in terms of non-numerical data, we first perform one-hot encoding, then use the embedding layer concatenating with financial data together to obtain the corporate embedding expression $x$ via
\begin{equation}
\label{eq1}
    x = Emb(c)
\end{equation}
where $x\in\mathbb{R}^d$. $d$ denotes the corporate embedding  size. Fig. \ref{C2G} illustrates the corporation-to-graph including three steps: self-outer product, activation function and graph construction.
\\

~\\
\noindent \textbf{Step1: Self-outer Product}

Above the embedding layer, we propose to use a self-outer product operation on $x$ to obtain the interaction map $\widetilde A$:
\begin{equation}
\label{eq2}
    \widetilde A=x\otimes x=xx^T
\end{equation}
where $\widetilde A$ is a $d\times d$ matrix.

This is the core design of C2G to ensure the effectiveness of feature interactions. We argue that using outer product is more advantageous in twofold: 1) it encodes more signals by accounting for the correlations between different features; 2) it is more meaningful than the original information in embedding without modeling any correlation. Recently, it has been shown that, it is particularly meaningful to model the interaction of feature explicitly, where as concatenation is sub-optimal \cite{beutel2018latent,he2018outer,he2017neural}.

~\\
\noindent \textbf{Step2: Activation Function}

Above the interaction map is an activation function, which targets at extracting useful signal from the interaction map. To be specific, it retains the important information and ignores the insignificant information. And it is subjected to design and can be abstracted as $A=g\left( \widetilde A \right)$, where $g$ denotes the activation function, and $A$ is the output matrix used for graph construction.
\begin{equation}
\label{eq3}
  A_{ij} = g\left( \widetilde A_{ij} \right)=
   \begin{cases} 
   0     &\quad \widetilde A_{ij} < r  \\   
    1    &\quad \widetilde  A_{ij} \geqslant  r   
   \end{cases}
\end{equation}
where $A_{ij}$ means the entry of matrix A that lies in the row number $i$ and column number $j$, $r$ is a adaptive parameter to get a connected graph. In other words, $A$ is the adjacent matrix of the graph.

~\\
\noindent \textbf{Step3: Graph Construction}

After getting the adjacent matrix $A$, we can construct this graph $\mathcal{G}=\left( \mathcal{V},\mathcal{E} \right)$. Let $\mathcal{V}=\{v_1,v_2,\cdots,v_d\}$, each graph has $d$ feature nodes. If $A_{ij}=1$, each edge $(v_i,v_j)\in \mathcal{E}$ means that a feature node $v_i$ has a strong relation with feature node $v_j$. Every node $v_i$ corresponds to an attribute $x_i \in \mathbb{R}$. Finally, the algorithm is formally presented in Algorithm \ref{alg:algorithm1}.

\begin{algorithm}[t]
	\caption{Corporation To Graph (G2G)}
	\label{alg:algorithm1}
	\KwIn{The set of corporations: $C$; Iteration magnitude $c$}
	\KwOut{The connected graph set $G$, $\mathcal{G}=\left( \mathcal{V},\mathcal{E} \right)\in G$ with node attribute $x$.}  
	\BlankLine
	Initialize empty graph set $G$
	
	\ForEach{\textnormal corporation $c$ in $C$}{
	    Obtain the corporate embedding expression $x$ via Equation \ref{eq1}\\
	    Get feature interaction map $\widetilde A$ by Equation \ref{eq2}\\
	    Initial threshold value $r=max\left(\widetilde A\right)$\\
	    \While{True}{
	        Compute adjacent matrix $A$ via Equation \ref{eq3} under threshold $r$\\
	        Construct graph $\mathcal{G}=\left( \mathcal{V},\mathcal{E} \right)$ with $A$\\
	        \eIf{$\mathcal{G}$ is a connected graph}{break;}{$r = r -c$;}
        }
    Add the graph $\mathcal{G}$ into graph set $G$
    }
 \textbf{return} graph set $G$

\end{algorithm}

	
	        

\subsection{Graph Feature Interaction Layer (GFIL)}
On the top of corporation to graph layer, we utilize graph attention network to simulate different importance feature interactions. Graph attention networks can automatically model feature interactions with attention mechanism. 

We can further stack more graph attention network layers to explore the high-order information, gathering the information propagated from the higher-hop neighbors. More formally, in the $l$-th GAT layer, for the node $v_i$ of graph $\mathcal{G}$ the update function is recursively formulated as follows:
\begin{equation}
    x_i^{(l)}=\alpha_{i,i}^{(l)}\Theta^{(l)} x_i^{(l-1)}+\sum_{j\in \mathcal{N}_{(i)}}\alpha_{i,j}^{(l)}\Theta^{(l)}x_j^{(l-1)}
\end{equation}
where the $\Theta^{(l)}\in \mathbb{R}^{d^{(l)}\times d^{(l-1)}}$ and $a\in \mathbb{R}^{2d^{(l)}}$ are GAT layer parameters, $x^{(l-1)}\in \mathbb{R}^{d\times d^{(l-1)}}$ is the input of GAT layer, the output $x^{(l)}\in \mathbb{R}^{d\times d^{(l)}}$ and the attention coefficients $\alpha_{i,j}^{(l)}$ are computed as:
\begin{equation}
    \alpha_{i,j}^{l}=\frac{\exp \left( LeakReLU\left(a^{T^{(l)}}\left[\Theta^{(l)} x_i^{(l-1)}\parallel \Theta^{(l)} x_j^{(l-1)}\right]\right)\right)}
    {\sum_{k\in \mathcal{N}(i)\cup \{i\}}\exp \left( LeakReLU\left(a^{T^{(l)}}\left[\Theta^{(l)} x_i^{(l-1)}\parallel \Theta^{(l)} x_k^{(l-1)}\right]\right)\right)}
\end{equation}

In the first layer of GFIL, the layer input is the corporate graph with features. In other words, $x^{(0)}=x\in \mathbb{R}^{d}$. Clearly, the high-order feature interactions are modeled into the representation learning process.

~\\
\noindent \textbf{Graph Pooling.} Besides local high-order feature interaction, graph pooling is designed after each GAT layers, in order to merge the global information. The pooling process can be formulated as:
\begin{equation}
    z^{(l)}=GraphPooling(x^{(l)})
\end{equation}
where $z^{(l)}\in \mathbb{R}^{d^{(l)}}$ is graph global information at $l$-th layer. In despite of some intricate methods, here we use some simple but effective mechanisms to implement it, such as max pooling, mean pooling.

\subsection{Credit Rating Layer (CRL)}
In GFIL layer, we get local high-order feature interactions through GAT layer and global graph information through pooling. After performing $L$ layers, we obtain multiple node representations, namely $\{x^{(0)},x^{(1)},\cdots,x^{(L)}\}$ and graph representations, namely $\{z^{(1)},z^{(2)},\cdots,z^{(L)}\}$. Inspired by ResNet and DenseNet, we aggregate local and global information firstly, then perform credit rating process by Multi-Layer Perceptron (MLP), which can be formulated as follows:
\begin{equation}
\begin{split}
    r_{local}& = concatenate(reshape(\{x^{(0)},x^{(1)},\cdots,x^{(L)}\}))\\
    r_{global} &= concatenate(\{z^{(1)},z^{(2)},\cdots,z^{(L)}\})
\end{split}
\end{equation}
where $r_{local}\in \mathbb{R}^{d(1+\sum_{l=1}^{L}d^{(l)})}$ and $r_{global}\in \mathbb{R}^{\sum_{l=1}^{L}d^{(l)}}$. Finally, the result of credit rating $\hat{y}$ is executed by MLP and softmax.

\begin{equation}
    \hat{y} = log\_softmax\left(MLP\left(\left[r_{local};r_{global}\right]\right)\right)
\end{equation}
where $\hat{y}\in \mathbb{R}^m$  denotes the probabilities of labels.

For training model, the loss function is defined as the cross-entropy of the prediction and the ground truth, it can be written as follows:
\begin{equation}
    \mathcal{L}(\hat{y})=-\sum\limits_{i=1}^{m}{{{y}_{i}}\log ({{{\hat{y}}}_{i}})+(1-{{y}_{i}})\log (1-{{{\hat{y}}}_{i}})}+\lambda {{\left\| \Delta  \right\|}^{2}}
\end{equation}
where $y$ denotes the one-hot encoding vector of ground truth item, $\lambda$ is parameter specific regularization hyperparameters to prevent overfitting, and the model parameters of CCR-GNN are $\Delta$ .

Finally,the Back-Propagation algorithm is performed to train the proposed CCR-GNN model.

\section{Experiments}\label{section:Experiments}
In this section, we describe the extensive experiments for evaluating the effectiveness of our proposed methods. We describe the datasets at first,   then present the experimental results of CCR-GNN compared with other baselines, which is the main task of this paper. 
\subsection{Data Set and Pre-processing}
The corporate credit dataset has been built based on the annual financial statements of Chinese listed companies and China Stock Market \& Accounting Research Database (CSMRA). The results of credit ratings are conducted by famous credit rating agency, including CCXI, China Lianhe Credit Rating (CLCR), etc. Real-world data is often noisy and incomplete. Therefore, the first step of any prediction problem to credit rating, in particular, is to clean data such that we maintain as much meaningful information as possible. Specifically, we delete features which miss most of value, and for features with a few missing values are filled in by the mean value. After data pre-processing which we use min-max normalization for numerical data and one-hot encoding for category data, we get 39 features and 9 rating labels: AAA, AA, A, BBB, BB, B, CCC, CC, C. The table \ref{tab1} will show details. Then Synthetic Minority Oversampling Technique (SMOTE) is conducted to perform data augmentation in terms of class-imbalance problem.

\begin{table}[htbp]
\centering
\caption{Data Set Description.}
\label{tab1}
\resizebox{1.0\columnwidth}{!}{
\setlength{\tabcolsep}{2.8mm}{
\begin{tabular}{|c|c|c|c|}
\hline
\textbf{Index} & \multicolumn{2}{c|}{\textbf{Criterion layer}}                                     & \textbf{Feature Name }           \\[2pt] \hline
1     & \multirow{12}{*}{Financial data} & \multirow{2}{*}{Profit Capability}    & Net Income              \\ \cline{1-1} \cline{4-4} 
…     &                                  &                                       & …                       \\ \cline{1-1} \cline{3-4} 
6     &                                  & \multirow{2}{*}{Operation Capability} & Inventory Turn Ratio    \\ \cline{1-1} \cline{4-4} 
…     &                                  &                                       & …                       \\ \cline{1-1} \cline{3-4} 
11    &                                  & \multirow{2}{*}{Growth Capability}    & Year-on-year Asset      \\ \cline{1-1} \cline{4-4} 
…     &                                  &                                       & …                       \\ \cline{1-1} \cline{3-4} 
18    &                                  & \multirow{2}{*}{Repayment Capability} & Liability To Asset      \\ \cline{1-1} \cline{4-4} 
…     &                                  &                                       & …                       \\ \cline{1-1} \cline{3-4} 
25    &                                  & \multirow{2}{*}{Cash Flow Capability} & Ebit To Interest        \\ \cline{1-1} \cline{4-4} 
…     &                                  &                                       & …                       \\ \cline{1-1} \cline{3-4} 
30    &                                  & \multirow{2}{*}{Dupont Identity}      & Dupont Return on Equity \\ \cline{1-1} \cline{4-4} 
…     &                                  &                                       & …                       \\ \hline
…     & \multicolumn{2}{c|}{\multirow{2}{*}{Non-Financial Data}}                 & …                       \\ \cline{1-1} \cline{4-4} 
39    & \multicolumn{2}{c|}{}                                                    & Tax Credit Rating       \\ \hline
\end{tabular}}}
\end{table}

\subsection{Comparison with Baseline Methods}
\noindent \textbf{Baseline Methods.}
We use the following widely used approaches in the financial field as baselines to highlight the effectiveness of our proposed methods:
\begin{itemize}
    \item[$\bullet$] \textbf{LR:} Logistic regression (LR) \cite{mcmahan2011follow} model for multi-label classification. We apply L2 normalization and follow the-regularized-leader (FTRL) for optimization.
    \item[$\bullet$] \textbf{SVM:} Support Vector Machine with linear kernel.
    \item[$\bullet$] \textbf{MLP:} Multi-Layer Perceptron, A simple neural network. We use 1000 hidden units in the experiments and ReLU for activation function.
    \item[$\bullet$] \textbf{GBDT:} Gradient Boosting Decision Tree \cite{ke2017lightgbm}, it is an iterative decision tree algorithm. The algorithm, which is a popular ensemble learning method for classification, consists of multiple decision trees, and the conclusions of all trees are added together to make the final answer.
    \item[$\bullet$] \textbf{Xgboost:} eXtreme Gradient Boosting \cite{chen2016xgboost}, a scalable machine learning system for tree boosting.
\end{itemize}
\noindent \textbf{Hyper-parameter Setup.}
In our experiments, we use three graph attention layers, attention channels $=\{8,64,9\}$ respectively. First two graph attention layers use Mean Pooling and the last is Max Pooling. All parameters are initialized using a Xavier uniform distribution with a mean of 0, and a standard deviation of 0.1. The Adam optimizer is exerted to optimize these parameters, where the initial learning rate is set to 0.001 and will decay by 0.0001 after every 3 training epochs. Moreover, the L2 penalty is set to 0.00001.

~\\
\noindent \textbf{Evaluation Metrics.}We adopt three commonly-used metrics for evaluation, including precision, recall and F1-score. In detail, we count the number of correct identification of positive labels as True Positives TP, incorrect identification of positive labels in False Positives FP, incorrect identification of positive labels in False Positives FP and incorrect identification of negative labels in False Negative FN. Then these metrics can be calculated as:
\begin{equation}
    \begin{split}
Recall&=\sum_{i=1}^m\frac{TP_i}{TP_i+FN_i}\\
Precision&=\sum_{i=1}^m\frac{TP_i}{TP_i+FP_i}\\
F1-Score&=\sum_{i=1}^m\frac{2\left(Precision_i * Recall_i\right)}{\left(Precision_i + Recall_i\right)}
\end{split}
\end{equation}
where $m$ is the number of labels, $Precision_i$ and $Recall_i$ are the metrics for $i$-th class respectively.

~\\
\noindent \textbf{Experiments Results.}
To demonstrate the overall performance of the proposed model, we compare it with other baseline models. The overall performance in terms of recall, accuracy and F1-score is shown in Table \ref{tab2}. The best results are highlighted as bold. The xgboost is second only to CCR-GNN. 
\begin{table}[htbp]
\centering
\caption{The Performance of CCR-GNN with other Baselines}
\label{tab2}
\setlength{\tabcolsep}{6mm}{
\begin{tabular}{cccc}
\toprule[1.5pt]
{\textbf{Model}} & { \textbf{Recall}} & {\textbf{Accuracy}} & { \textbf{F1-score}} \\ \hline
LR      & 0.76250 & 0.80970  & 0.81946 \\
SVM     & 0.83750 & 0.89247  & 0.88961 \\
MLP     & 0.91406 & 0.93568  & 0.93254 \\
GBDT    & 0.91875 & 0.92647  & 0.9187 \\
xgboost & \underline{0.92343} & \underline{0.94225} & \underline{0.94133} \\ \hline
CCR-GNN & \textbf{0.93437} & \textbf{0.95012}  & \textbf{0.95177} \\
\bottomrule[1.5pt]
\end{tabular}}
\end{table}

According to the experiments, it is obvious that the proposed CCR-GNN method achieves the best performance on real dataset (Chinese public-listed corporate rating dataset). By stacking multiple graph attention layers, CCR-GNN is capable of exploring the high-order feature interactions in an explict way. This verifies the effectiveness of the proposed method.

\section{Conclusions}\label{section:conclusion}
In this work, we develop a novel graph neural network for corporate credit rating named CCR-GNN. Corporation-To-Graph is proposed for each corporation to build a graph according to the relations between features. By utilizing the power of GAT, CCR-GNN can capture both local and global information. Besides, extensive experiments on Chinese public-listed corporate rating dataset are conducted to demonstrate the effectiveness of our model. Therefore, CCR-GNN provides an effective tool for financial regulators to control potential credit risk.

\section{Acknowledgements}
This work is jointly supported by National Natural Science Foundation of China (62071017) and Major Projects in Tianjin Binhai New District (BHXQKJXM-PT-ZKZNSBY-2018001).

\bibliographystyle{splncs04}
\bibliography{CCRGNN}
\end{document}